\documentclass[conference,anonymous]{IEEEtran}
\IEEEoverridecommandlockouts
\usepackage{multirow}
\usepackage{cite}
\usepackage{amsmath,amssymb,amsfonts}
\usepackage{algorithmic}
\usepackage{graphicx}
\usepackage{textcomp}
\usepackage{xcolor}
\usepackage[colorlinks,
            urlcolor=blue,
            linkcolor=blue,       
            anchorcolor=blue,  
            citecolor=blue,        
            ]{hyperref}

\def\BibTeX{{\rm B\kern-.05em{\sc i\kern-.025em b}\kern-.08em
    T\kern-.1667em\lower.7ex\hbox{E}\kern-.125emX}}
\begin{document}

\title{A Multi-Stage Goal-Driven Network for Pedestrian Trajectory Prediction\\
\thanks{This research was funded by Fujian NSF (2022J011112) and the Research Project of Fashu Foundation (MFK23001).}
}
\author{\IEEEauthorblockN{Xiuen Wu$^{1,2}$, Tao Wang$^{1,\ast}$, Yuanzheng Cai$^{1}$, Lingyu Liang$^{3}$, George Papageorgiou$^{4}$}
\IEEEauthorblockA{$^{1}$Fujian Provincial Key Laboratory of Information Processing and Intelligent Control, Minjiang University, Fuzhou, China. \\
$^{2}$College of Computer and Data Science, Fuzhou University, Fuzhou, China. \\
$^{3}$School of Electronic and Information Engineering, South China University of Technology, Guangzhou, China.\\
$^{4}$SYSTEMA Research Center, European University Cyprus, Nicosia, Cyprus.}
$^{*}$Corresponding author: Tao Wang, E-mail: twang@mju.edu.cn.}

\maketitle

\begin{abstract}
Pedestrian trajectory prediction plays a pivotal role in ensuring the safety and efficiency of various applications, including autonomous vehicles and traffic management systems. This paper proposes a novel method for pedestrian trajectory prediction, called multi-stage goal-driven network (MGNet).  Diverging from prior approaches relying on stepwise recursive prediction and the singular forecasting of a long-term goal, MGNet directs trajectory generation by forecasting intermediate stage goals, thereby reducing prediction errors. The network comprises three main components: a conditional variational autoencoder (CVAE), an attention module, and a multi-stage goal evaluator.
Trajectories are encoded using conditional variational autoencoders to acquire knowledge about  the approximate distribution of pedestrians' future trajectories, and combined with an attention mechanism to capture the temporal dependency between trajectory sequences. The pivotal module is the multi-stage goal evaluator, which utilizes the encoded feature vectors to predict intermediate goals, effectively minimizing cumulative errors in the recursive inference process.
The effectiveness of MGNet is demonstrated through comprehensive experiments on the JAAD and PIE datasets. Comparative evaluations against state-of-the-art algorithms reveal significant performance improvements achieved by our proposed method.

\end{abstract}


\section{Introduction}

With the rapid advancement of autonomous driving technologies, the research and application of pedestrian trajectory prediction have increasingly attracted widespread attention. Pedestrians are an important part of the urban transportation system, and predicting their behavior patterns and trajectories has a crucial impact on traffic flow analysis, autonomous driving, intelligent monitoring, and urban planning\cite{okuda2014survey,cocsar2016toward,yurtsever2020survey,sighencea2021review}. 
In interactive environments, the comprehension and prediction of pedestrian trajectories play a pivotal role in ensuring the safe navigation of autonomous driving systems. For instance, the precision of predicting pedestrians' future trajectories is imperative for self-driving cars to avert collisions and uphold road safety. There is also a need to strategize secure and socially-aware paths while issuing alerts for unusual movements.
Moreover, in intelligent monitoring, predicting pedestrian trajectories proves beneficial for detecting abnormal behaviors and potential safety risks. Real-time analysis of pedestrian trajectories enables prompt detection of safety hazards, including situations such as pedestrians in non-traffic areas or abnormal gatherings of pedestrian groups. Subsequently, appropriate measures can be implemented to prevent accidents.

\begin{figure}[t!]
\centerline{\includegraphics[width=1\linewidth]{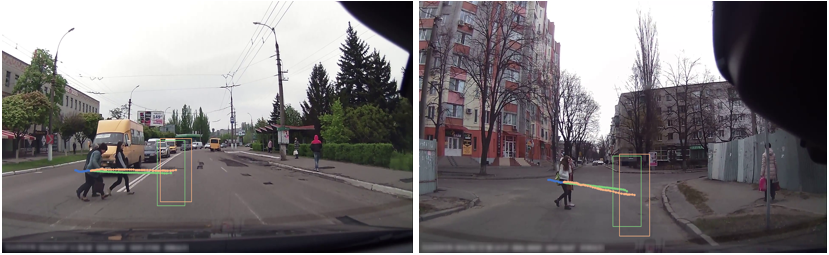}}
\caption{Example of trajectory prediction results. The blue line represents the past trajectory, the orange line represents the actual future trajectory, and the green line represents the predicted trajectory.}
\label{fig1}
\end{figure}

In recent years, the swift advancement of deep learning technology, notably with the introduction of recurrent neural networks (RNN) and long short-term memory networks (LSTM), has substantially elevated the accuracy of pedestrian trajectory prediction. This progress has spawned a large amount of related research, leveraging a series of technologies such as attention mechanisms \cite{huang2019stgat,sadeghian2019sophie,yu2020spatio,chiara2022goal}, graph neural networks \cite{sun2020recursive,mohamed2020social,shi2021sgcn}, generative models \cite{gupta2018social,liang2021temporal,lee2017desire}, and goal-driven networks \cite{albrecht2021interpretable,mangalam2020not,deo2020trajectory,dendorfer2020goal,mangalam2021goals} to better capture complex spatio-temporal relationships and understand pedestrian movement patterns, thereby achieving more accurate predictions of pedestrians' future trajectories.\par

Although significant research has been conducted in recent years, the majority of studies have focused on aerial views or fixed cameras in video surveillance, rather than cameras installed on autonomous vehicles. The primary distinction is that the latter must account for ego-motion, thereby complicating pedestrian trajectory prediction from an egocentric perspective due to the influence of ego-motion captured by in-vehicle cameras. Most of the work \cite{rasouli2019pie,makansi2020multimodal,chandra2019traphic,neumann2021pedestrian,huynh2023online} based on the egocentric perspective utilizes additional information such as ego-motion, semantic intention, image features or social interaction to some extent. 
However, in recent years, certain goal-driven models \cite{yao2021bitrap,wang2022stepwise} solely utilize observed trajectories as input and direct the generation of future trajectories by predicting either a long-term goal or step-by-step goals. Their performance surpasses that of many methods requiring additional features. Thus, we initiated a study on enhancing trajectory prediction performance through a goal-driven approach without the reliance on additional features. Diverging from existing goal-driven models, which typically estimate only the final destination or employ stepwise recursion, we introduce a multi-stage goal-driven network (MGNet). We posit that predicting stage goals can more effectively steer the forward recursive reasoning of the trajectory, consequently reducing cumulative errors in the reasoning process and enhancing long-term trajectory prediction. In real-life scenarios, pedestrians often need to plan a series of goals to guide their travel direction when aiming to reach a destination.  This multi-stage goal is more detailed than a single goal and more accurately represents the pedestrian's movement intention. Based on this assumption, we employ a conditional variational autoencoder (CVAE) augmented with an attention mechanism as the encoder. The CVAE learns the distribution of future trajectories conditioned on observed past trajectories through stochastic latent variables, while the attention mechanism captures complex temporal dependencies to better predict multi-stage goals. Then experiments show that estimating multi-stage goals provides superior guidance for predicting future trajectories.


The primary contributions of this work can be summarized into the following four parts:
\begin{itemize}
    \item We propose a novel pedestrian trajectory prediction network (MGNet) designed to mitigate errors in recursive prediction by anticipating multiple staged goals.
    \item We employ the attention mechanism to integrate the trajectory distribution features acquired by CVAE, enhancing the prediction of staged goals in the future trajectory and consequently achieving further performance improvement.
    \item We crafted a multi-stage goal evaluator characterized by a double-layer structure. This evaluator utilizes the goal features from the upper layer to inform the output of subsequent layers of goals.
    \item Superior results are attained in the JAAD\cite{kotseruba2016joint} and PIE\cite{rasouli2019pie} datasets when compared to the current state-of-the-art models.
\end{itemize}
\par
The remainder of the paper is structured as follows: Section II offers a thorough review of related works. Section III delineates the essential components of the proposed methods. Section IV outlines the experimental details and furnishes both qualitative and quantitative comparisons. Finally, Section V summarizes the findings and engages in a discussion of future works.
\begin{figure*}[htbp]
\centerline{\includegraphics[width=1\linewidth]{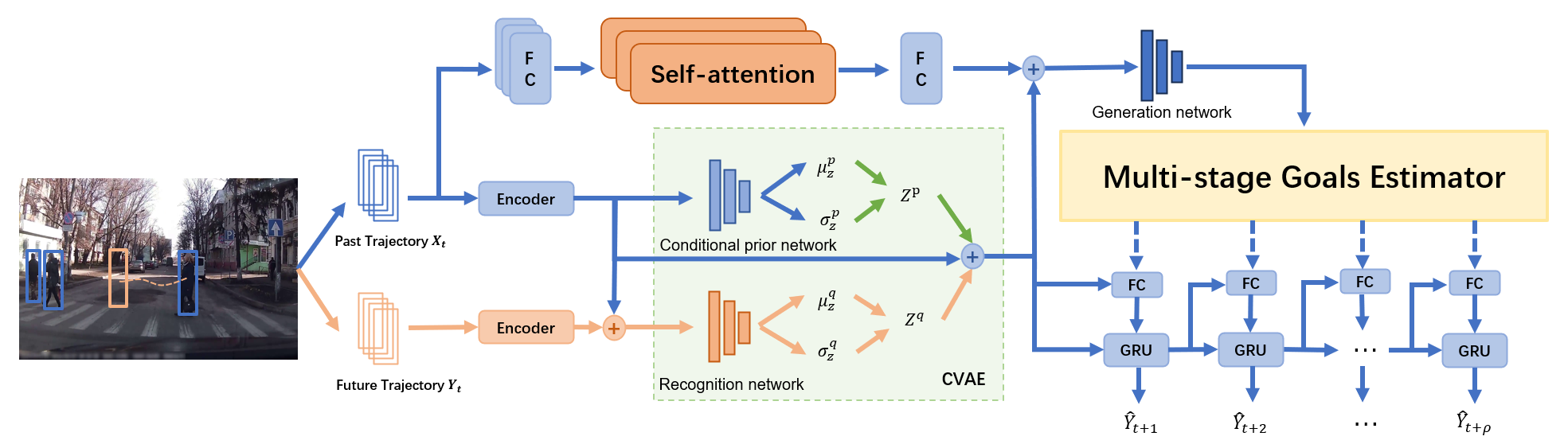}}
\caption{
An overview of our MGNet Architecture. arrows in orange, green, and blue denote connections during training, inference, and both training and inference, respectively.}
\label{fig3}
\end{figure*}
\begin{figure}[htbp]
\centerline{\includegraphics[width=1\linewidth]{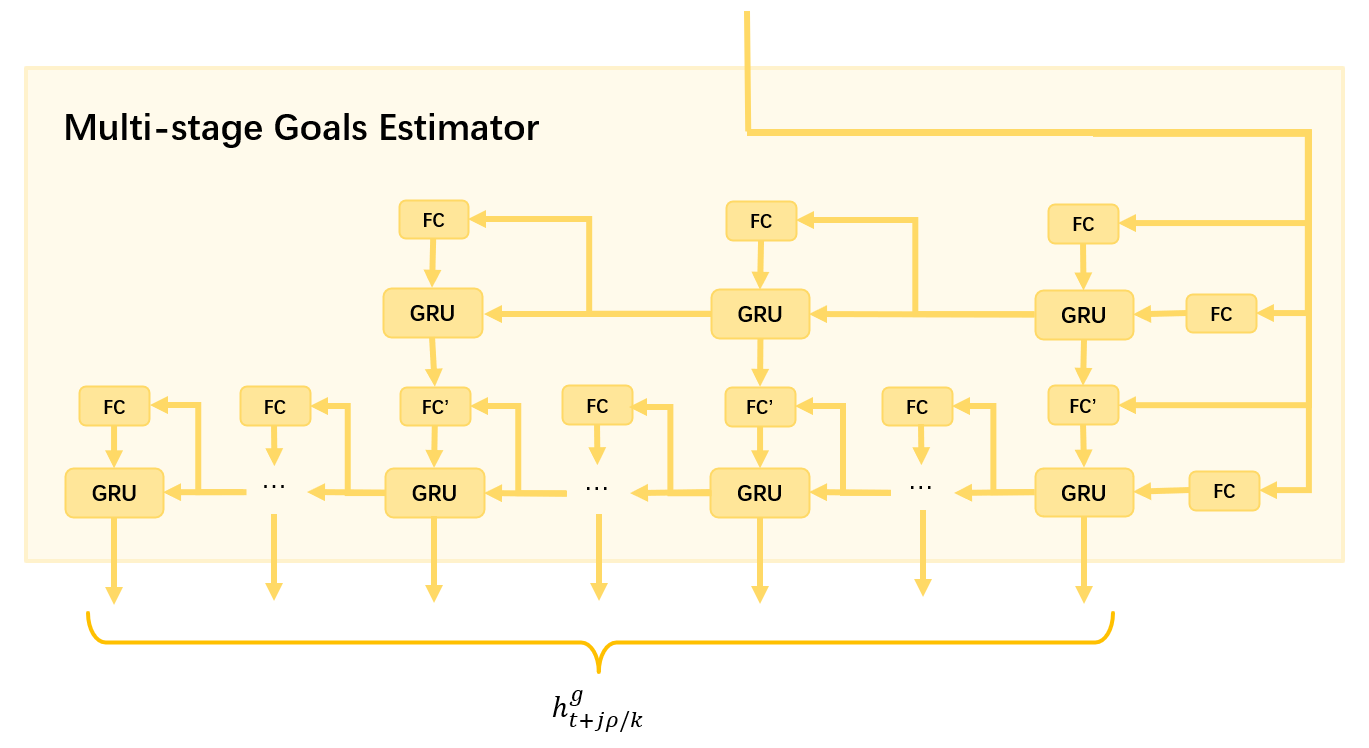}}
\caption{Detailed structure of Multi-Stage Goal Evaluator. It adopts a double-layer structure, and uses the stage goal features predicted by the upper layer to guide the generation of lower-level stage goals, and outputs several stage goal features.}
\label{fig4}
\end{figure}
\section{Related Work}

\subsection{Trajectory Prediction in Dynamic Video Scenes}
The egocentric camera perspective is frequently considered the most natural viewpoint for observing the surrounding environment of an ego-vehicle. However, it poses additional challenges due to its limited field of view and ego-motion. Several studies have addressed these challenges by transforming the perspective into a bird’s-eye view using 3D sensors~\cite{chai2019multipath,gao2020vectornet,salzmann2020trajectron++,song2022learning,zhou2022hivt,aydemir2023adapt}. While this method is feasible, it is susceptible to measurement errors and multimodal data processing issues, especially with LiDAR and stereo sensors. \par
Therefore, there are also many studies conducted directly under the egocentric view. Bhattacharyya et al.~\cite{bhattacharyya2018long} employed Bayesian Long Short-Term Memory (LSTM) networks to model observation uncertainty and integrated them with ego-motion to predict the distribution of potential future positions.  Yagi et al.~\cite{yagi2018future} utilized information such as pose, locations, scales, and past ego-motion to predict the future trajectory of a person. Chandra et al.~\cite{chandra2019traphic} models the interrelationships between nearby heterogeneous objects to predict trajectories. Yao et al.~\cite{yao2019egocentric} proposed a novel multi-stream RNN encoder-decoder model that independently captures both object location and appearance. Makansi et al.~\cite{makansi2020multimodal} estimate a reachability prior for objects based on the semantic map and project this information into the future. Diverging from the aforementioned approaches, we abstain from employing additional features such as self-motion, semantic intent, and image characteristics. Instead, we exclusively utilize observed past trajectories as input to accomplish pedestrian trajectory prediction from an egocentric perspective.

\subsection{Attention-based Methods for Trajectory Prediction}
In recent years, more and more studies have shown the effectiveness of attention mechanisms in trajectory prediction. Huang et al. \cite{huang2019stgat} integrated a Graph Attention Network (GAT) with Long Short-Term Memory (LSTM) to model pedestrian motion. Sadeghian et al. \cite{sadeghian2019sophie} combine attention mechanisms with social interactions and physical information between pedestrians to generate future trajectories. Yu et al. \cite{yu2020spatio} proposed a spatio-temporal graph transformer framework that exclusively employs the attention mechanism to address trajectory prediction. chiara et al. \cite{chiara2022goal} introduce a straightforward yet effective attention-based recurrent architecture designed to handle temporal dependencies. nayakanti et al. \cite{nayakanti2023wayformer} proposed an attention-based scene encoder-decoder capable of fusing one or more modalities across temporal and spatial dimensions. In our work, we integrate the attention mechanism with the generative model CVAE to enhance the generation of stage goals by capturing more intricate temporal dependencies.

\subsection{Goal-Driven Methods for Trajectory Prediction}
Several works leverage goal prediction to improve prediction accuracy. Mangalam et al.\cite{mangalam2020not} assist in long-range multi-modal trajectory prediction by inferring distant trajectory endpoints. Rhinehart et al.~\cite{rhinehart2019precog} introduce a generative multi-agent forecasting method capable of conditioning on agent goals and modeling the relationships between individual agent goals. Zhao et al. \cite{zhao2021tnt} predict an agent's potential future goal states by encoding its interactions with the environment and other agents, and then generate a trajectory state sequence conditioned on the goal. Yao et al. \cite{yao2021bitrap} employs a bidirectional decoder on the predicted goal to enhance long-term trajectory prediction. Wang et al. \cite{wang2022stepwise} predict a series of stepwise goals at various temporal scales and integrate them into both encoders and decoders for trajectory prediction. Building upon the existing work, we have extended beyond the prediction of a single long-term goal or stepwise goals to guide trajectory generation. We have designed a multi-stage goal evaluator to predict multiple stage goals within future trajectories. By leveraging stage goal features, our aim is to reduce cumulative errors in recursive inference, thereby generating more accurate trajectories.

\section{Proposed Method}
\subsection{Problem Formulation}

As shown in Fig. \ref{fig1}, the purpose of trajectory prediction is to anticipate the future position sequence of the target in a scene based on the observed past position sequence. At time step t,  we use $\mathbf{X}_t = [\mathbf{X}^1_t,\mathbf{X}^2_t,...,\mathbf{X}^n_t]$ to represent the past trajectories of n pedestrians. Where $\mathbf{X}^i_t = [X^i_{t-\tau+1},X^i_{t-\tau+2},...,X^i_t]$ is the set of observation positions of pedestrian i in the past $\tau$ frames. For $X^i_t = (x^i_t,y^i_t,w^i_t,h^i_t)$, it represents the position and size of a bounding box, where $(x^i_t,y^i_t)$ denotes the center coordinates, and $(w^i_t,h^i_t)$ represents the width and height of the bounding box, respectively. Given $\mathbf{X}_t$, a predictor can be employed to forecast the future trajectories $\mathbf{\hat{Y}}_t = [\mathbf{\hat{Y}}^1_t,\mathbf{\hat{Y}}^2_t,...,\mathbf{\hat{Y}}^n_t ]$ for all n pedestrians, where $\mathbf{\hat{Y}}^i_t = [\hat{Y}^i_{t+1},\hat{Y}^i_{t+2},...,\hat{Y}^i_{t+\rho}]$ represents the position of pedestrian i over the next $\rho$ frames. 
The definition of $\hat{Y}^i_{t+1} = (\hat{x}^i_{t+1},\hat{y}^i_{t+1},\hat{w}^i_{t+1},\hat{h}^i_{t+1})$ is similar to $X^i_t$ and denotes a bounding box. We aim to design a model such that the predicted future trajectory $\mathbf{\hat{Y}}_t$ closely aligns with the actual future trajectory $\mathbf{Y}_t$. The definitions of $\mathbf{Y}_t$ and $\mathbf{\hat{Y}}_t$ are similar. Furthermore, we use $\mathbf{G}_t = [\mathbf{G}^1_t,\mathbf{G}^2_t,...,\mathbf{G}^n_t]$ to denote a series of stage goals for n pedestrians output by the multi-stage goal evaluator. Then, $\mathbf{G}^i_t = [g^i_{t+\rho/k},g^i_{t+2\rho/k},...,g^i_{t+\rho}]$ represents the stage goal position of pedestrian i in the upcoming $\rho$ frames, where $k \in [1, \rho]$ denotes the division of the trajectory for the next $\rho$ frames into $k$ stage goals.

\subsection{Overview}
The model we propose is primarily structured around three key modules: the conditional variational autoencoder, the attention module, and the multi-stage goal evaluator. The encoding part of the model can be roughly divided into two sections. In the first part, to capture more intricate temporal dependencies, we utilize the parallel self-attention mechanism in Transformers to encode the past trajectory $X_t$, resulting in the feature vector $h_A$. In the second part, the past trajectory $X_t$ needs to be encoded through a Gated Recurrent Unit (GRU) encoder to obtain the encoded feature vector $h_X$. During the training phase, the ground truth goal $Y_t$ is encoded by another GRU encoder and concatenated with ht to generate the feature vector $h_Y$. Subsequently, the feature vectors $h_X$ and $h_Y$ are respectively input into the conditional prior network and recognition network within the CVAE module to learn the distribution of future trajectories. The resulting output features are concatenated with the feature vectors $h_A$ and $h_X$, and fed into the generation network and the forward recursive inference. The generation network then derives the feature vectors necessary for multi-stage goal evaluation. Finally, in the decoding stage, the multi-stage goal evaluator outputs hidden features for several stages of goals and incrementally incorporates them into the forward recursive inference process, resulting in the final future trajectory. The overall architecture of the proposed model is illustrated in Fig. \ref{fig3}.

\subsection{Temporal Attention}
In recent years, Transformer\cite{vaswani2017attention} has achieved great success in time series data prediction. It pioneered self-attention, which can assign different attention weights to different positions in the input sequence, greatly improving the accuracy of time series data prediction. At the same time, multiple self-attention heads are also introduced to learn multiple sets of different attention weights, which enhances the model's ability to model input sequences. In our work, we use the encoding part of Transformer. Given a past observed trajectory sequence $\mathbf{X}_t$, three different vectors are obtained through linear transformation: $Q$(query), $K$ (key) and $V$(value). The attention weights are then calculated using these three vectors with the following formula:
\begin{align}
 \text{Attention}(Q, K, V) = \text{softmax}(\frac{{Q K^T}}{\sqrt{d_k}})V
\end{align}
where $d_k$ is the dimension of $K$, used as the normalization factor. Based on the aforementioned formula, the multi-head attention (k heads) conducts multiple parallel attention calculations on distinct segments of the input sequence, its formula is expressed as:
\begin{align}
 \text{MultiHead}(Q, K, V) = \text{Concat}(head_1,...,head_m)f_O \\
 \text{where $head_j$} = \text{Attention}(Q_j, K_j, V_j)
\end{align}
where $f_O$ is a fully connected layer that merges the outputs of $m$ attention heads. The final embedding $h_A$ is generated by a fully connected layers.


\subsection{Conditional Variational Autoencoder}
We also used a conditional variational autoencoder(CVAE) to encode the pedestrian's past trajectory sequence to derive the latent variable $Z$, so as to learn and generate an approximate distribution of future trajectories. 
With reference to \cite{lee2017desire,mangalam2020not,yao2021bitrap}, Our CVAE consists of the following modules: 1) Recognition network $Q_\phi(Z^q|\mathbf{X}_t,\mathbf{Y}_t)$, responsible for capturing the correlation between variable $Z$ and the actual trajectory y. 2) Conditional prior network $P_\theta(Z^p|\mathbf{X}_t)$, tasked with modeling the latent variable $Z$ based on past observed trajectories. 3) Generation network $P_\omega(h_G|\mathbf{X}_t,Z,h_A)$, responsible for encoding input features and generating multi-stage goals. Here, $\phi,\theta,\omega$ denote the parameters of corresponding networks. The three mentioned networks consist of three-layer multi-layer perceptrons.
Different from previous work, we augment the generation network by incorporating latent features $h_A$ encoded by multi-head self-attention, so that it can learn richer temporal dependence information.  \par
During the training stage, the past trajectory $\mathbf{X}_t$ and the ground truth future trajectory $\mathbf{Y}_t$ are initially encoded using distinct gated-recurrent unit  encoders to obtain the feature vectors $h_X$ and $h_Y$, respectively.
To capture dependence information between past trajectories and ground truth future trajectories, we input the feature vectors $h_X$ and $h_Y$ into the recognition network to predict the distribution mean $\mu^q_z$ and standard deviation $\sigma^q_z$ of future trajectories. The conditional prior network assumes that only $h_X$ is utilized to predict the distribution mean $\mu^p_z$ and standard deviation $\sigma^p_z$, without knowledge of the ground truth future trajectory.
Next, we sample $Z^q$ from $\mathcal{N}(\mu^q_z,\sigma^q_z)$, combine it with $h_X$ and $h_A$, and ultimately utilize the generation network to generate the hidden features $h_G$ required by the multi-stage goal evaluator. During the testing phase, the ground truth future trajectories are not available.  So unlike the training phase, $Z^p$ is sampled from $\mathcal{N}(\mu^p_z,\sigma^p_z)$ to generate $h_G$.

\subsection{Multi-Stage Goal Estimator}
As shown in Fig. \ref{fig4}, the structure diagram of the multi-stage goal evaluator. It employs a double-layer architecture, proceeding from coarse to fine in a top-down manner. The features of upper-layer predicted stage goals guide the generation of lower-layer stage goals. Specifically, the evaluation device comprises a double-layer reverse RNN. Each layer is initialized with the feature vector $h_G$ as input and initialization, and generates the hidden feature of several stages of goals from time $t+\rho$ to $t+1$. In the first layer, the entire trajectory is divided into three stages for prediction, while in the second layer, further refinement divides it into more stages. The specific number of segments can be adjusted based on the final model's effectiveness, with subsequent experimental sections discussing this matter accordingly. At the same time step, the coarse-grained goal features of the upper layer are concatenated with the hidden features of the GRU in the lower layer to predict the fine-grained goal feature $h^g_{t+j \rho/k}$, which is then used as the output of the evaluator. Hence, we can define the output of the multi-stage target evaluator as:
\begin{align}
 h^g_{t+j \rho/k} = f_{MSGE}(ReLU(W_fh_G + b_f))
\end{align}
Where $j=\{1,2,3,...,k\}$ denotes the index of the $j\text{-th}$ stage goal, $\rho$ signifies the number of time steps to be predicted, and $k\in[1,\rho]$ represents dividing the future trajectory into $k$ stage goals. $f_{MSGE}$ represents the entire multi-goal evaluator.
Meanwhile, to compute the stage goal loss, it is fed into a fully connected layer to generate the stage goal $G_t$.\par
During the final decoding stage, the stage target output by the evaluator at the same time step is concatenated with the hidden state of forward recursive inference to predict the final trajectory waypoint at that time.  It's worth noting that due to the variable number of stage target features output by the multi-objective evaluator, the connection between the evaluator and forward recursive reasoning in Fig. \ref{fig3} is represented by a dotted line.

\begin{table*}[htbp]
    \renewcommand{\arraystretch}{1.3}
  \centering
  \caption{ Quantitative results on JAAD and PIE datasets. Lower values are better. }
  \setlength{\tabcolsep}{4.3mm}{
    \begin{tabular}{c|cccccc|cccccc}
    \hline
    \multicolumn{1}{c|}{\multirow{2}[4]{*}{Method}} & \multicolumn{6}{c}{JAAD}              & \multicolumn{6}{c}{PIE} \\
\cline{2-13}          & \multicolumn{4}{c}{MSE} & $C_{MSE}$  & $CF_{MSE}$  & \multicolumn{4}{c}{MSE} & $C_{MSE}$  & $CF_{MSE}$ \\
          & \multicolumn{4}{c}{( 0.5 / 1.0 / 1.5s )} & (1.5s) & (1.5s) & \multicolumn{4}{c}{( 0.5 / 1.0 / 1.5s )} & (1.5s) & (1.5s) \\
    \hline
    {Linear \cite{rasouli2019pie}} & \multicolumn{4}{c}{ 223 / 857 / 2303 } & 1565  & 6111  & \multicolumn{4}{c}{123 / 477 / 1365} & 950  & 3983 \\
    {LSTM \cite{rasouli2019pie}}  & \multicolumn{4}{c}{289 / 569 / 1558} & 1473  & 5766  & \multicolumn{4}{c}{172 / 330 / 911} & 837  & 3352 \\
    {B-LSTM \cite{bhattacharyya2018long}} & \multicolumn{4}{c}{159 / 539 / 1535} & 1447  & 5615  & \multicolumn{4}{c}{101 / 296 / 855} & 811  & 3259 \\
    {FOL-X \cite{yao2019egocentric}} & \multicolumn{4}{c}{147 / 484 / 1374} & 1290  & 4924  & \multicolumn{4}{c}{47 / 183 / 584} & 546  & 2303 \\
    {PIEtraj \cite{rasouli2019pie}} & \multicolumn{4}{c}{110 / 399 / 1248} & 1183  & 4780  & \multicolumn{4}{c}{58 / 200 / 636} & 596  & 2477 \\
    {PIEfull \cite{rasouli2019pie}} & \multicolumn{4}{c}{-} & -  & -  & \multicolumn{4}{c}{- / - / 559} & 520  & 2162 \\
    {BiTraP-D \cite{yao2021bitrap}} & \multicolumn{4}{c}{93 / 378 / 1206} & 1105  & 4565  & \multicolumn{4}{c}{41 / 161 / 511} & 481  & 1949 \\
    \hline
    {MGNet} & \multicolumn{4}{c}{\textbf{87} / \textbf{353} / \textbf{1132}} & \textbf{1079}  & \textbf{4452}  & \multicolumn{4}{c}{\textbf{37} / \textbf{145} / \textbf{474}} & \textbf{445}  & \textbf{1906} \\
    \hline
    \end{tabular}}%
    
  \label{tab1}%
\end{table*}%

\subsection{Loss Functions}
The model's overall loss function comprises three components: trajectory prediction loss, goals loss, and The KL-divergence (KLD) loss. The trajectory prediction loss quantifies the error between the predicted trajectory $\mathbf{X}_t$ by the model and the ground truth future trajectory $\mathbf{Y}_t$. The goals loss measures the error between the staged goals $\mathbf{G}_t$ predicted by the multi-stage goal estimator and the corresponding real goals $\mathbf{Y}^g_t$. For the aforementioned two losses, we utilize the L2 norm for calculation. The formulas for the two losses are expressed as follows:

\begin{equation}
    L_{pred} = \|\mathbf{X}_t - \mathbf{Y}_t\|_2
\end{equation}
\begin{equation}
    L_{goals} = \|\mathbf{G}_t - \mathbf{Y}^g_t\|_2
\end{equation}

The KLD loss captures the difference between the distribution $\mathcal{N}(\mu^q_z,\sigma^q_z)$ output by the recognition network of the CVAE module and the distribution $\mathcal{N}(\mu^p_z,\sigma^p_z)$ output by the prior network in the same module. The total losses, including KLD losses, are as follows:

\begin{equation}
    L_{total} =  L_{pred} + L_{goals} +  KLD(Z^q \,||\, Z^p)
\end{equation}



\begin{table*}[htbp]
    \renewcommand{\arraystretch}{1.3}
  \centering
  \caption{ Exploration study of our model on JAAD and PIE datasets. Lower values are better.}
  \setlength{\tabcolsep}{4.3mm}{
    \begin{tabular}{c|cccccc|cccccc}
    \hline
    \multicolumn{1}{c|}{\multirow{2}[4]{*}{Goal}} & \multicolumn{6}{c}{JAAD}              & \multicolumn{6}{c}{PIE} \\
\cline{2-13}          & \multicolumn{4}{c}{MSE} & $C_{MSE}$  & $CF_{MSE}$  & \multicolumn{4}{c}{MSE} & $C_{MSE}$  & $CF_{MSE}$ \\
          & \multicolumn{4}{c}{( 0.5 / 1.0 / 1.5s )} & (1.5s) & (1.5s) & \multicolumn{4}{c}{( 0.5 / 1.0 / 1.5s )} & (1.5s) & (1.5s) \\
    \hline
    {1} & \multicolumn{4}{c}{ 92 / 376 / 1212 } & 1147  & 4683  & \multicolumn{4}{c}{45 / 164 / 525} & 493  & 2190 \\
    {3}  & \multicolumn{4}{c}{90 / 365 / 1157} & 1101  & 4537  & \multicolumn{4}{c}{39 / 150 / 490} & 460  & 1973 \\
    {9} & \multicolumn{4}{c}{90 / 368 / 1179} & 1132  & 4638  & \multicolumn{4}{c}{\textbf{37} / \textbf{145} / \textbf{474}} & \textbf{445}  & \textbf{1906} \\
    {15} & \multicolumn{4}{c}{\textbf{87} / \textbf{353} / \textbf{1132}} & \textbf{1079}  & \textbf{4452}  & \multicolumn{4}{c}{40 / 150 / 484} & 454  & 1975 \\
    {45} & \multicolumn{4}{c}{89 / 359 / 1159} & 1106  & 4565  & \multicolumn{4}{c}{41 / 156 / 503} & 473  & 2054 \\
    \hline
    \end{tabular}}%
  \label{tab2}%
\end{table*}%

\section{Experiments}

\subsection{Datasets}
For our experiments, we assess the effectiveness of our framework by conducting evaluations on the JAAD\cite{kotseruba2016joint} and PIE\cite{rasouli2019pie} datasets, specifically for the prediction of pedestrian trajectories from ego-centric perspectives.  JAAD includes 2,800 pedestrian trajectories captured from dash cameras and annotated at 30Hz. PIE comprises 1,842 pedestrian trajectories, also annotated at 30Hz, featuring longer trajectories and more comprehensive annotations, including semantic intention, ego-motion, and neighboring objects.
Following \cite{rasouli2019pie}, we divided the datasets into training, testing, and validation sets with ratios of 50\%, 40\%, and 10\%, respectively. Additionally, we use an observation length of 0.5 seconds to predict future trajectories of lengths 0.5, 1.0, and 1.5 seconds.

\subsection{Implementation Details}

We conducted experiments on a desktop server running Ubuntu 20.04 OS equipped with a 4.00GHz Intel Core i9-9900KS CPU, 64GB RAM, and a single NVIDIA GeForce RTX 3090 GPU. In our model architecture, Gated Recurrent Units (GRUs) serve as the backbone for both the encoder and decoder, each configured with a hidden size of 256.
For the attention module, We utilize an embedding dimension of size 32 for spatial coordinates and employ a single transformer encoder layer with 8 multi-head attention heads. 
We utilize the Adam optimizer with an initial learning rate of 0.001, dynamically adjusted based on the validation loss. We use the ReLU as the activation function, and to address overfitting, we implement batch normalization and dropout in our model. The optimization is performed end-to-end with a batch size of 128, and the training process concludes after 100 epochs.

\subsection{Evaluation Metrics}

In this paper, we mainly utilize the mean squared error (MSE) between the upper left and lower right corners of the bounding box to evaluate the performance of our proposed approach. The specific evaluation formula is as follows:

\begin{align}
\text{MSE} = \frac{1}{n} \sum_{i=1}^{n} (y_i - \hat{y}_i)^2
\end{align}
where n is the number of predicted samples, $y_i$ is the ground truth, and $\hat{y}_i$ is the predicted value of the model.  \par
In addition, we also calculated the center mean squared error ($C_{MSE}$) and the center final mean squared error ($CF_{MSE}$) for result evaluation. $C_{MSE}$ can measure the accuracy of the entire trajectory, while $CF_{MSE}$ only measures the accuracy of the endpoints of the trajectory. The calculation formulas for both are similar to MSE, but note that their calculations are based on the bounding box centroid. All metrics are measured in pixels.

\begin{table}[htbp]
    \renewcommand{\arraystretch}{1.3}
  \centering
  \caption{ Ablation study of our model on JAAD datasets. Lower values are better. BL: the baseline model, AT: the attention module, ES: the multi-stage goal evaluator.}
  \setlength{\tabcolsep}{3.0mm}{
    \begin{tabular}{ccccccccc}
    \hline
    \multirow{2}[1]{*}{BL} & \multirow{2}[1]{*}{AT} & \multirow{2}[1]{*}{ES}  & \multicolumn{4}{c}{MSE} & $C_{MSE}$  & $CF_{MSE}$   \\
    \hfill & \hfill & \hfill & \multicolumn{4}{c}{( 0.5 / 1.0 / 1.5s )} & (1.5s) & (1.5s)  \\
    \hline
    
     \checkmark & \hfill & \hfill & \multicolumn{4}{c}{ 94 / 383 / 1231 } & 1190  & 4734  \\
     \checkmark & \checkmark & \hfill & \multicolumn{4}{c}{ 92 / 376 / 1212 } & 1147  & 4683  \\
     \checkmark & \hfill & \checkmark & \multicolumn{4}{c}{ 89 / 362 / 1164 } & 1113  & 4573  \\
     \checkmark & \checkmark & \checkmark & \multicolumn{4}{c}{ 87 / 353 / 1132 } & 1079  & 4452  \\
    \hline
    \end{tabular}}
  \label{tab:addlabel}
\end{table}


\subsection{Results}
 \textbf{Quantitative Comparison.} To ensure the reliability of the experimental results, we derived the model's output from the average of three experiments. As depicted in Table \ref{tab1}, we conducted a comparative analysis of our model against the current state-of-the-art algorithms on the JAAD and PIE datasets, achieving superior results across various indicators. Particularly noteworthy, compared to the BiTraP model, MGNet demonstrated an average performance increase of 6.3$\%$ and 8.9$\%$ under the MSE indicator on the JAAD and PIE datasets, respectively. These results  underscore the pivotal role of guiding trajectory generation through predicting multi-stage goals in enhancing prediction accuracy.\par
\textbf{Exploration Study} We examine the impact on results by adjusting the number of stage goal features output by the multi-stage goal estimator, with the results presented in Table \ref{tab2}. In the table, the output one goal signifies that the model does not utilize a multi-stage goal estimator and solely predicts a long-term goal to guide trajectory generation. The output of three goals represents a single-layer structure used in the multi-stage goal estimator, segmenting the future trajectory into three stages to guide trajectory generation. Experimental data indicates that on the JAAD dataset, the best results are achieved when 15 goal features are output, whereas on the PIE dataset, optimal results are obtained with 9 goal features. This emphasizes that the optimal number of output stage goal features is dataset-dependent and requires further adjustment. Additionally, it is observed that when the multi-stage goal evaluator is not used, that is, when only one goal feature is output, the results are the least favorable, providing further evidence of the efficacy of the multi-stage goal evaluator. \par
\textbf{Ablation Study.} In the JAAD dataset, we systematically remove various components from the model to assess their impact on the experimental results. The findings are presented in Table \ref{tab:addlabel}. "BL" in the table denotes the baseline, which solely utilizes CVAE encoding and outputs a single long-term goal to guide forward recursive inference for trajectory prediction. "AT" represents the attention module, while "ES" stands for the multi-stage goal evaluator. It is evident from the table that incorporating either the attention module or the multi-stage goal evaluator positively influences the results, with the multi-stage goal evaluator demonstrating the most significant improvement. Ultimately, the optimal results are achieved by combining the attention module and the multi-stage goal evaluator.

\begin{figure*}[htbp]
\centerline{\includegraphics[width=1\linewidth]{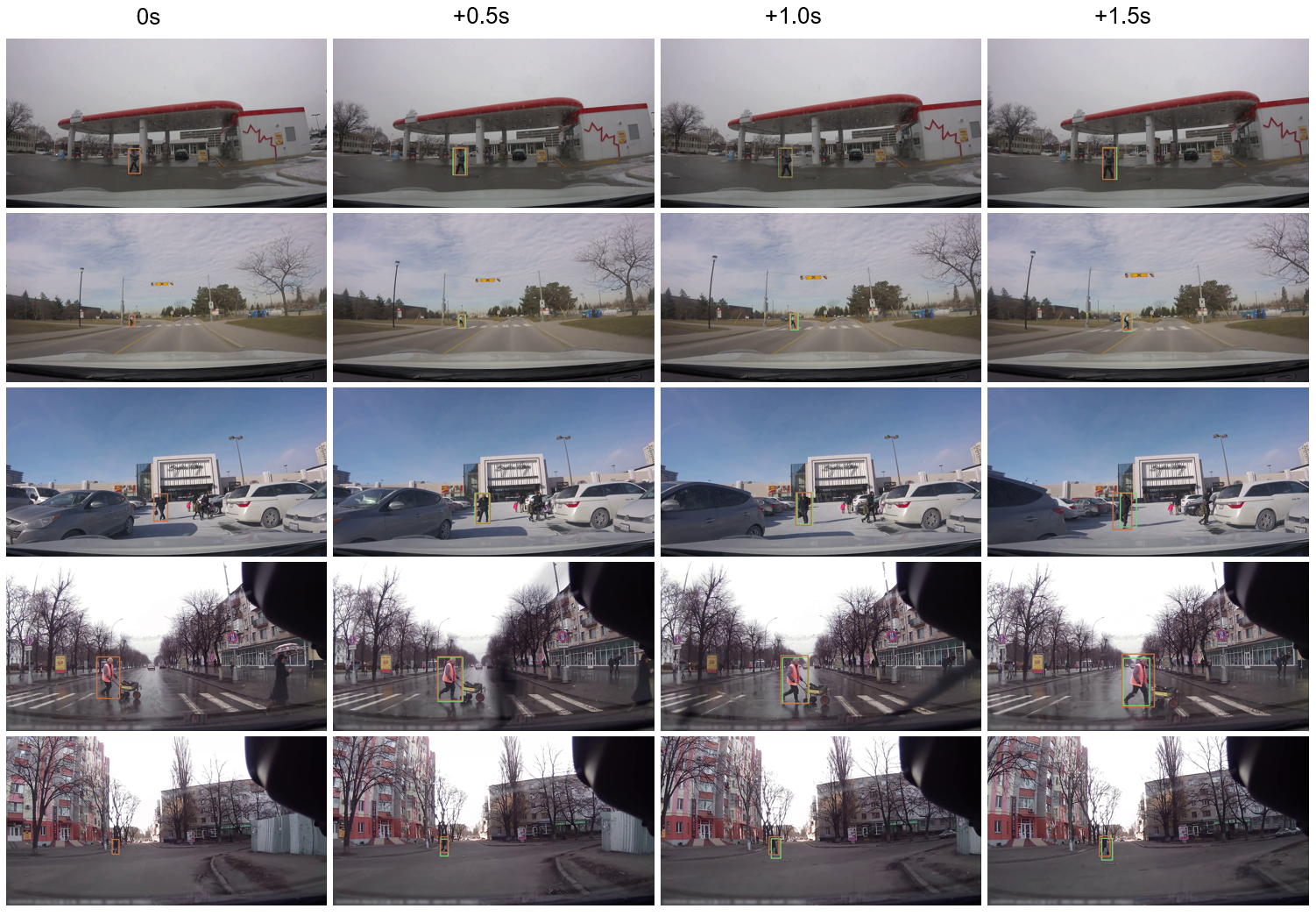}}
\caption{Qualitative results on trajectory prediction on the JAAD dataset. The ground truth is orange and the predictions is green. Best viewed in color.}
\label{fig2}
\end{figure*}

 \textbf{Qualitative Results.} As shown in Fig. \ref{fig2}, we visualize the prediction bounding boxes and ground truth at several key time nodes (0.5s, 1.0s, 1.5s) to qualitatively demonstrate the prediction performance. The samples are extracted from the JAAD dataset, covering five representative scenes in daily life: 1) gas station, 2) main road in town, 3) parking lot, 4) crosswalk, and 5) intersection. In the depicted figure, our predicted bounding boxes closely align with the ground truth. Specifically, within the next 1.0 second, the predicted bounding boxes exhibits nearly identical location and size to the ground truth. At the subsequent 1.5 second, although the position and size of the predicted bounding boxes deviate from the actual situation, the overall position remains roughly consistent.



\section{Conclusion}
In this work, we propose a new goal-driven network model (MGNet) for pedestrian trajectory prediction. Leveraging the generative model CVAE and an attention module, the model encodes trajectories and subsequently generates multiple stage goals through a multi-stage goal evaluator to guide future trajectory generation. Unlike most existing goal-driven models that exclusively estimate the final destination or distant goals, our approach posits that predicted stage goals can more effectively guide the forward recursive inference of the trajectory, thus reducing cumulative errors in the inference process. Experimental results substantiate the effectiveness of the proposed model compared to state-of-the-art methods. In future work, we aim to explore the construction of a trajectory memory model to enhance the guidance of stage goal generation and further improve model performance.

\bibliographystyle{IEEEtran}
\bibliography{IEEEabrv,ref}


\end{document}